\documentclass[accepted]{uai2023} % after acceptance, for a revised
\usepackage[american]{babel}
\usepackage{natbib} % has a nice set of citation styles and commands
\usepackage{mathtools} % amsmath with fixes and additions
\usepackage{booktabs} % commands to create good-looking tables
\usepackage{tikz} % nice language for creating drawings and diagrams
\usepackage{amsmath}
\usepackage{amsfonts,amssymb}
\usepackage{calligra}
\usepackage{algorithm,algorithmicx,algpseudocode}
\usepackage{multirow}
\usepackage{bm}
\usepackage{booktabs}
\usepackage{verbatim}
\usepackage{subfigure}
\usepackage[graphicx]{realboxes}
\usepackage{algorithm}% http://ctan.org/pkg/algorithm
\usepackage{algpseudocode}% http://ctan.org/pkg/algorithmicx

\errorcontextlines\maxdimen

\makeatletter
\newcommand*{\algrule}[1][\algorithmicindent]{\makebox[#1][l]{\hspace*{.5em}\thealgruleextra\vrule height \thealgruleheight depth \thealgruledepth}}%
\newcommand*{\thealgruleextra}{}
\newcommand*{\thealgruleheight}{.75\baselineskip}
\newcommand*{\thealgruledepth}{.25\baselineskip}

\newcount\ALG@printindent@tempcnta
\def\ALG@printindent{%
	\ifnum \theALG@nested>0% is there anything to print
	\ifx\ALG@text\ALG@x@notext% is this an end group without any text?
	\else
	\unskip
	\addvspace{-1pt}% FUDGE to make the rules line up
	\ALG@printindent@tempcnta=1
	\loop
	\algrule[\csname ALG@ind@\the\ALG@printindent@tempcnta\endcsname]%
	\advance \ALG@printindent@tempcnta 1
	\ifnum \ALG@printindent@tempcnta<\numexpr\theALG@nested+1\relax% can't do <=, so add one to RHS and use < instead
	\repeat
	\fi
	\fi
}%
\usepackage{etoolbox}
\patchcmd{\ALG@doentity}{\noindent\hskip\ALG@tlm}{\ALG@printindent}{}{\errmessage{failed to patch}}
\makeatother

\newbox\statebox
\newcommand{\myState}[1]{%
	\setbox\statebox=\vbox{#1}%
	\edef\thealgruleheight{\dimexpr \the\ht\statebox+1pt\relax}%
	\edef\thealgruledepth{\dimexpr \the\dp\statebox+1pt\relax}%
	\ifdim\thealgruleheight<.75\baselineskip
	\def\thealgruleheight{\dimexpr .75\baselineskip+1pt\relax}%
	\fi
	\ifdim\thealgruledepth<.25\baselineskip
	\def\thealgruledepth{\dimexpr .25\baselineskip+1pt\relax}%
	\fi
	\State #1%
	\def\thealgruleheight{\dimexpr .75\baselineskip+1pt\relax}%
	\def\thealgruledepth{\dimexpr .25\baselineskip+1pt\relax}%
}
\title{Improvable Gap Balancing for Multi-Task Learning}

\author[1,3]{\href{mailto:<yanqidai@ruc.edu.cn>}{Yanqi~Dai}}
\author[2,3]{Nanyi~Fei}
\author[1,3]{Zhiwu~Lu}
\affil[1]{%
    Gaoling School of Artificial Intelligence\\
    Renmin University of China\\
    Beijing, China
}
\affil[2]{%
    School of Information\\
    Renmin University of China\\
    Beijing, China
}

\affil[3]{
    Beijing Key Laboratory of Big Data Management and Analysis Methods\\ 
    Beijing, China
}
  
\begin{document}

\maketitle

\begin{abstract}
In multi-task learning (MTL), gradient balancing has recently attracted more research interest than loss balancing since it often leads to better performance. However, loss balancing is much more efficient than gradient balancing, and thus it is still worth further exploration in MTL. Note that prior studies typically ignore that there exist varying improvable gaps across multiple tasks, where the improvable gap per task is defined as the distance between the current training progress and desired final training progress. Therefore, after loss balancing, the performance imbalance still arises in many cases. In this paper, following the loss balancing framework, we propose two novel improvable gap balancing (IGB) algorithms for MTL: one takes a simple heuristic, and the other (for the first time) deploys deep reinforcement learning for MTL. Particularly, instead of directly balancing the losses in MTL, both algorithms choose to dynamically assign task weights for improvable gap balancing. 
Moreover, we combine IGB and gradient balancing to show the complementarity between the two types of algorithms.
Extensive experiments on two benchmark datasets demonstrate that our IGB algorithms lead to the best results in MTL via loss balancing and achieve further improvements when combined with gradient balancing.
Code is available at \href{https://github.com/YanqiDai/IGB4MTL}{https://github.com/YanqiDai/IGB4MTL}.
\end{abstract}

\section{Introduction}
\label{sec:intro}

Multi-task learning (MTL) is to jointly train a single model that can perform multiple tasks \citep{caruana1998multitask, ruder2017overview, zhang2021survey, vandenhende2021multi}.
Compared with single-task learning (STL), MTL has two remarkable advantages: 1) the model typically has a smaller size and higher learning efficiency by sharing parameters across tasks \citep{misra2016cross, yang2020multi, vandenhende2021multi}, and 2) the performance on some tasks can be further improved due to the correlation between different tasks \citep{swersky2013multi}. 
Therefore, MTL has been widely used in real-world application scenarios such as recommendation systems \citep{zhao2019recommending} and automatic driving \citep{chowdhuri2019multinet}.
%, and robotic control \citep{kalashnikov2021mt}.

Note that the largest challenge in MTL is the seesaw phenomenon \citep{tang2020progressive}: joint training often results in better performance on some tasks but worse performance on others. To overcome this challenge, many optimization methods have been proposed for MTL: one is loss balancing that directly assigns different weights to the losses of multiple tasks according to a variety of criteria \citep{kendall2018multi, liu2019end, lin2022reasonable}; the other is gradient balancing that first calculates the task gradients and then aggregates them in different ways \citep{sener2018multi, liu2021conflict, navon2022multi}. 
Since gradient balancing typically performs better than loss balancing in MTL, it has attracted more research interest recently. However, the training cost of gradient balancing algorithms is significantly higher than that of most loss balancing algorithms, especially when there are significantly more tasks in MTL \citep{kurin2022defense}.

\begin{figure}[t]
 \centering
 \includegraphics[width=0.99\linewidth]{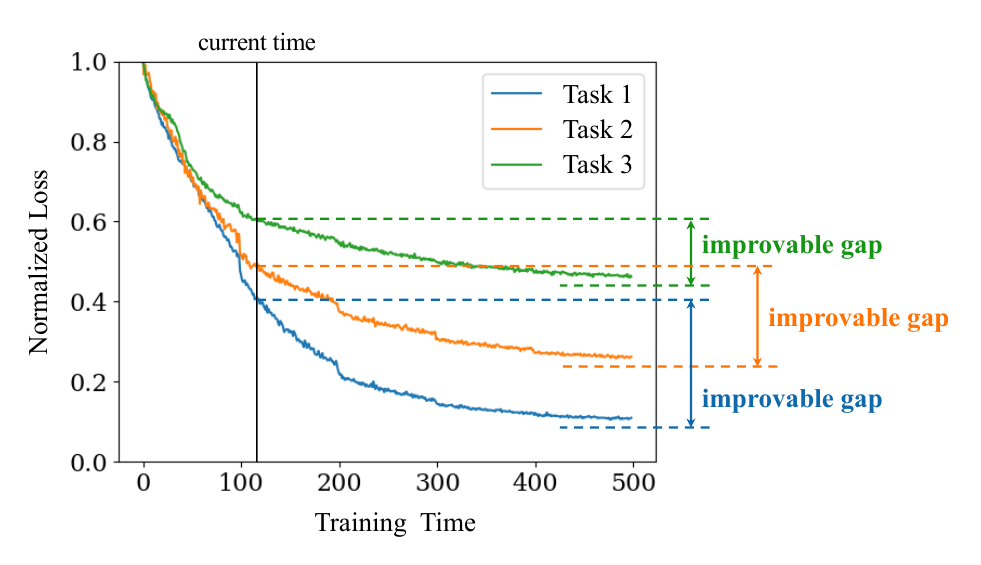}
 \caption{Schematic illustration of the improvable gaps for three different tasks in MTL. The improvable gap per task is defined as the distance between the current training progress and desired final training progress.}
 \label{fig:ig}
\end{figure}

Although most of prior studies focus on improving the performance of MTL by proposing more advancing methods, we emphasize that \textbf{improving the learning efficiency is also of the primary research significance in MTL}. Therefore, loss balancing and gradient balancing are both worth further study, and the trade-off between efficiency and performance can be taken according to practical requirements.
In this paper, we focus on further improving the performance of existing loss balancing methods. Specifically, within the loss balancing framework, we propose two novel improvable gap balancing algorithms, where the improvable gap per task is defined as the distance between the current training progress and desired final training progress. To define the improvable gap, we have to first represent the training progress as the loss decline normalized by the average loss across the current training time (from the beginning of training). As shown in Figure~\ref{fig:ig}, the normalized task losses tend to converge to nonzero values due to the limited training data and model capacity. Importantly, different tasks in MTL tend to have different convergence patterns (thus have different improvable gaps at the current training time). However, existing algorithms rarely notice that tasks trained together usually have different improvable gaps, which results in that some tasks are fully trained (or even making MTL outperform STL on these tasks), while others are still underfitted. Therefore, our main idea is to \textbf{dynamically assign task weights for improvable gap balancing (IGB)}, instead of loss balancing that has been widely used in MTL.

We propose the first algorithm IGBv1 to balance improvable gaps through a simple heuristic by uniformly defining the ideal loss as 0 for each task. 
Moreover, loss balancing can be interpreted as the maximum cumulative loss decline in the future by assigning weights through sequential decisions, which is consistent with the application scenario of reinforcement learning \citep{kaelbling1996reinforcement}. Therefore, we propose another algorithm IGBv2 to jointly minimize all improvable gaps through deep reinforcement learning (DRL) \citep{arulkumaran2017deep}. Additionally, we combine IGB and gradient balancing to show the complementarity between the two types of algorithms.

In summary, our main contributions are three-fold:\\
\textbf{(1)} We propose two novel loss balancing algorithms to dynamically balancing the improvable gaps in MTL. To our best knowledge, we are the first to apply DRL to MTL.\\
\textbf{(2)} Extensive experiments on two benchmark datasets demonstrate that our IGB algorithms perform the best in loss balancing and yield further improvements when combined with gradient balancing.\\
\textbf{(3)} We rethink the significance of loss balancing in terms of learning efficiency as well as its complementarity with gradient balancing.

\section{Related Work}
\label{sec:related}

%\subsection{Multi-Task Learning}
\noindent\textbf{Multi-Task Learning.}~~
Multi-task learning (MTL) research is broadly divided into two categories: one is to learn the correlation between tasks through model structures \citep{misra2016cross, ma2018modeling, liu2019end}, and the other is to balance the joint training process of all tasks through optimization algorithms \citep{kendall2018multi, lin2022reasonable, sener2018multi, liu2021towards, navon2022multi}. Our research is primarily concerned with the latter, which can be categorized into two types: loss balancing and gradient balancing.

Loss balancing directly updates the model by adding up or averaging the weighted losses after assigning task weights according to a variety of criteria.
The training time of most loss balancing algorithms is nearly the same as that of STL, since the input data, such as losses of the current batch, is low-dimensional.
The most common method for loss balancing is Equal Weighting (EW), which directly minimizes the sum of task losses.
Besides, various criteria are considered in prior studies. For example,
\citet{kendall2018multi} measured task uncertainty through learnable parameters; 
\citet{guo2018dynamic} estimated task difficulty based on key performance;
\citet{liu2019end} considered change rate of loss; 
\citet{lin2022reasonable} randomly assigned task weights;
\citet{ye2021multi} focused on metamodel validation performance.

Gradient balancing, on the other hand, first calculates task gradients separately and then updates the model by aggregating task gradients in different ways.
The performance of gradient balancing is typically better than that of loss balancing, since it can deal with gradient conflicts \citep{yu2020gradient} directly at the gradient level. For example, 
\citet{chen2018gradnorm} normalized task gradients to learn each task at a similar rate; 
\citet{sener2018multi} regarded gradient aggregation as a multi-objective optimization problem;
\citet{liu2021conflict} added the condition of minimum average loss to \citet{sener2018multi}; 
\citet{yu2020gradient} projected task gradients onto the normal planes of conflicting gradients;
\citet{chen2020just} randomly dropped some task gradients;
\citet{liu2021towards} aimed to make the aggregate gradient contribute equally to each task;
\citet{navon2022multi} considered gradient aggregation as a Nash bargaining game.
However, due to multiple backpropagations and high-dimensional gradient aggregation, gradient balancing often requires greater training time than STL, which severely limits its learning efficiency in practice.
Additionally, some works also explored the combination of loss balancing and gradient balancing, using loss balancing to update the task-specific parameters and combining loss balancing with gradient balancing to update the task-shared parameters \citep{liu2021towards, lin2022reasonable, liu2022auto}.

%\subsection{Reinforcement Learning}
\noindent\textbf{Reinforcement Learning.}~~
%\noindent\textbf{Soft Actor-Critic.}~~
Reinforcement learning (RL) is an interactive machine learning decision-making method for streaming data to maximize the expected return \citep{kaelbling1996reinforcement}. It is commonly modeled using a Markov decision process, which assumes that the future state is independent of the past state given the present state \citep{watkins1989learning}. In other words, let $s_i$ be the state at time $i$, the state $s_t$ is Markovian if and only if
\begin{equation}
	\Pr(s_{t+1} | s_t) = \Pr(s_{t+1} | s_1, s_2, \cdots, s_t).
\end{equation}

Soft Actor-Critic (SAC) \citep{haarnoja2018soft} is an off-policy actor-critic algorithm based on the maximum entropy RL framework, which is chosen as the loss weighting component in our IGBv2 algorithm.
The Actor-Critic algorithm \citep{peters2008reinforcement} combines policy-based RL with value-based RL. The actor generates actions and interacts with the environment, while the critic evaluates the performance of the actor and directs the actions in the subsequent stage. 
This algorithm is more efficient than policy gradient methods because it can be updated at each step.
The replay buffer \citep{mnih2013playing} in SAC is a classic off-policy mechanism where the agent learned and the agent interacting with the environment is different. It stores historical data for training, which is composed of four elements: the state $s_t$, the action $a_t$, the reward $r_t$ and the next state $s_{t+1}$.
The off-policy mechanism is beneficial to improve sample efficiency and reduce training instability caused by time series data.
The maximum entropy reinforcement learning \citep{ziebart2008maximum} changes the optimization objective to maximize both the expected return and the expected entropy of the policy. 
It increases the randomness of the policy, indicating that the probability distribution of the action is much wider.
Note that more randomness is proved beneficial to improve performance in MTL \citep{lin2022reasonable, chen2020just}.

In the context of RL, MTL can facilitate the transfer of knowledge between tasks, which has been shown to improve the performance of RL agents in various domains. For example, \citet{chen2021multi} trained a MTL agent for autonomic optical networks, effectively expediting the training processes and improving the overall service throughput. Our work differs from this approach in that we, for the first time, apply reinforcement learning to solve general MTL optimization problems.

\section{Methodology}
\label{sec:method}

In this section, we first provide a scale-invariant loss balancing paradigm. Based on this paradigm, we then describe our IGB algorithms. Finally, we give the fusion paradigm to combine loss balancing and gradient balancing.

\subsection{Scale-Invariant Loss Balancing}
\label{sec:si}

The objective of existing loss balancing algorithms is to minimize the weighted sum or average of task losses \citep{lin2021closer}. 
However, if task losses are on different scales, the model update is probably dominated by a specific task.

To solve the scale difference problem of task losses, \citet{navon2022multi} introduces a scale-invariant objective $\sum_{i=1}^n \log(L_i)$, where $n$ is the number of tasks and $L_i$ is the $i$th task loss. 
%The objective is scale-invariant due to the logarithmic operation, which can extract any constant coefficients outside and turn them into addition terms.
Inspired by this, we propose a scale-invariant loss balancing paradigm referred to as SI. The total loss of SI is defined as:
\begin{equation}\label{eq:si}
	L_{total} = \sum_{i=1}^n \lambda_{l,i} \log(L_i),
\end{equation}
where $\lambda_{l,i}$ is the $i$th task weight assigned by a specific loss balancing algorithm. 
In this paper, both of our IGB algorithms are designed based on the SI paradigm.

\subsection{Improvable Gap Balancing}
\label{sec:IGB}

We propose two loss balancing algorithms: IGBv1 directly balances improvable gaps across tasks estimated by a simple heuristic, while IGBv2 jointly minimizes all improvable gaps through a DRL model. 

\subsubsection{IGBv1}
\label{sec:IGBv1}

\begin{algorithm}[t]
	\renewcommand{\algorithmicrequire}{\textbf{Input:}}
	\renewcommand{\algorithmicensure}{\textbf{Output:}}
	\caption{Training iteration of IGBv1}  
	\label{alg:IGBv1}
	\begin{algorithmic}[1]
		\Require task number $n$, current batch losses $\bm{L}$, current epoch $ce$, current batch $cb$, batch number of one epoch $bn$, learning rate $\eta$, task-shared parameters $\theta$, task-specific parameters $\left\lbrace \psi_i \right\rbrace^n_{i=1}$
		\Ensure updated task-shared parameters $\theta'$, updated task-specific parameters $\left\lbrace\psi'_i\right\rbrace^n_{i=1}$
		
		\If {$ce=2$ \textbf{and} $cb=bn$}
		\State Assign $\bm{L_{base}}$ to the average losses across every batch in the current epoch of all tasks;
		\EndIf
		
		\If{$ce \leq 2$}
		\State $\bm{\lambda_{v1}} = \bm{1} \in \mathbb{R}^n$;
		\Else
		\State $\bm{\lambda_{v1}} = n \times \mathrm{softmax}(\bm{L}/\bm{L_{base}})$;
		\EndIf
		
		\State $\theta' = \theta - \eta \nabla_\theta \sum^n_{i=1}\lambda_{v1, i} \log(L_i)$;
		\For{i = 1 \textbf{to} n}
		\State $\psi_i' = \psi_i - \eta \nabla_{\psi_i} \lambda_{v1, i} \log(L_i)$;
		\EndFor
	\end{algorithmic}
\end{algorithm}

In the ideal situation, the MTL model can completely fit each training data, which satisfies $L_i = 0, \forall i$. Therefore, we assume that the desired final training losses of all tasks are 0, so that the improvable gaps can be directly represented by the normalized losses of the current batch, where the normalization is to deal with different task loss scales.

\begin{figure*}[t!]
    \centering
    \includegraphics[width=0.95\linewidth]{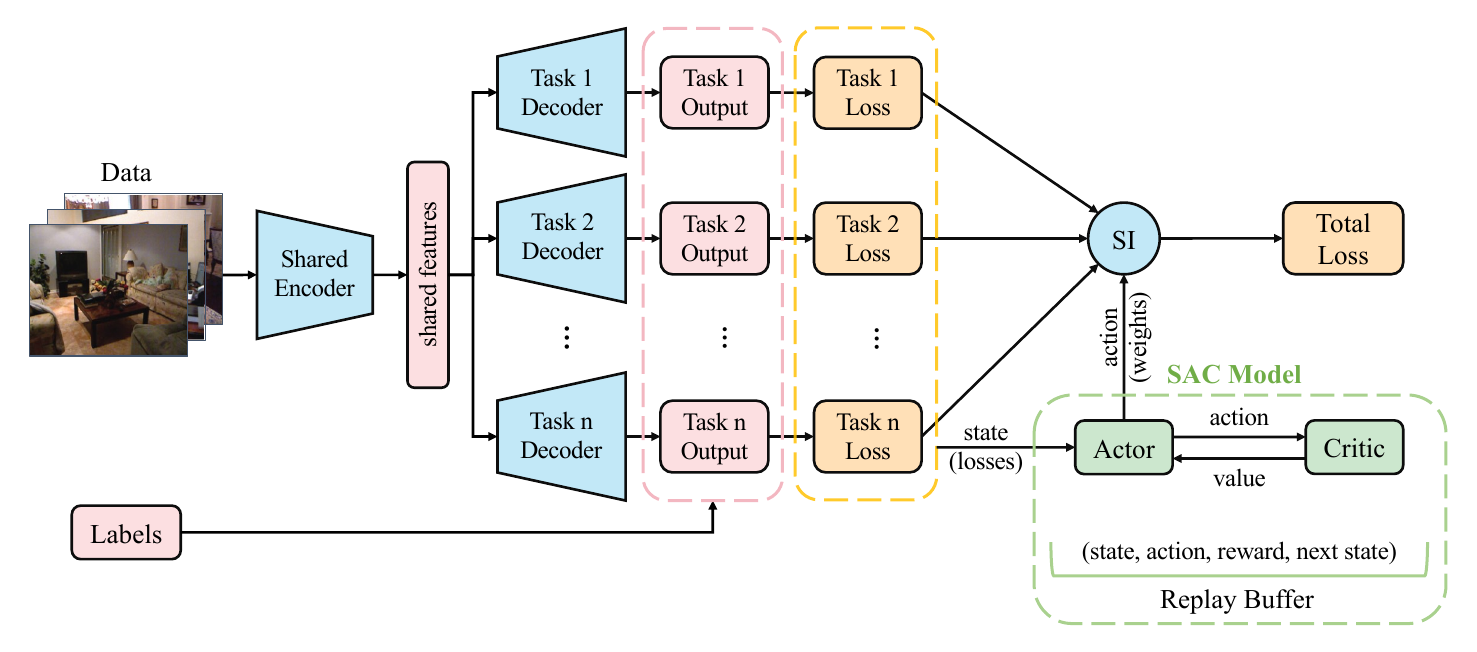}
    \caption{The architecture overview of our proposed IGBv2 algorithm by applying DRL to MTL.}
    \label{fig:IGBv2}
\end{figure*}

To normalize the training losses, we calculate the average losses across every batch in the second epoch of all tasks as $\bm{L_{base}} = [L_{base, 1}, L_{base, 2}, \ldots, L_{base,n}]$, since the losses of the first epoch may be too large to accurately represent the loss scales if the MTL model is randomly initialized.
In the first two epochs of training, the task weights are assigned to constant 1.
Then in the subsequent training process, the task weights are calculated as:
\begin{equation}
	\bm{\lambda_{v1}} = n \times \mathrm{softmax}\left(\frac{\bm{L}}{\bm{L_{base}}}\right), 
\end{equation}
where $\bm{L} = \left[L_1, \cdots, L_n\right]$ is the current batch losses of all tasks, and $\bm{L / L_{base}}$ denotes the element-wise division of the two vectors. 
By this way, we dynamically assign more weights to tasks which have more improvable gaps.
The training process of IGBv1 is summarized in Algorithm~\ref{alg:IGBv1}, where we initially set both the current epoch $cp$ and the current batch $cb$ to $1$.

\subsubsection{IGBv2}
\label{sec:IGBv2}

In practice, it is difficult for the MTL model to satisfy $L_i = 0, \forall i$ at the end of training. The reason is that existing deep learning methods cannot fully fit the training data distribution, and the minimal loss on the training data usually indicates overfitting. Therefore the estimation of the improvable gap in IGBv1 is inaccurate, and we present IGBv2 which can adaptively balance improvable gaps through DRL.
 
Note that assigning task weights for loss balancing can be interpreted as sequential decisions for maximizing the cumulative loss declines of tasks in the future, while the loss declines can be regarded as the feedback reward for task weights, which is consistent with the application scenario of RL.
Inspired by this, we propose deploying a DRL model to assign task weights, with the loss declines of the MTL model as the reward for DRL. 

We choose Soft Actor-Critic (SAC) \citep{haarnoja2018soft} as the DRL model to guide the training of the MTL model, which is an off-policy actor-critic algorithm based on the maximum entropy RL framework. 
As interpreted in Figure~\ref{fig:IGBv2}, the task weights are assigned by the SAC model with the losses of the MTL model as input, and the MTL model and the SAC model are alternately trained.
The reasons for choosing SAC are as follows:
1) The actor-critic structure \citep{peters2008reinforcement} allows the SAC model to be updated at each step, enabling IGBv2 to assign the current optimal weights in time;
2) The replay buffer \citep{mnih2013playing} in SAC is a commonly used off-policy mechanism, which can improve sample efficiency by training the SAC model on random historical data. Because the MTL model can only be trained once, the training data available for our SAC model is much less than that for typical DRL applications, making sample efficiency crucial;
3) The maximum entropy mechanism \citep{ziebart2008maximum} can increase the randomness of the action, thereby increasing the likelihood of discovering the global optimum of the MTL model.

\begin{algorithm}[t]
	\renewcommand{\algorithmicrequire}{\textbf{Input:}}
	\renewcommand{\algorithmicensure}{\textbf{Output:}}
	\caption{Training iteration of IGBv2}  
	\label{alg:IGBv2}
	\begin{algorithmic}[1]
		\Require task number $n$, current batch losses $\bm{L}$, current epoch $ce$, current batch $cb$, batch number of one epoch $bn$, SAC model $sac$, replay buffer $buffer$, start epoch to update the SAC model $update\_e$, start epoch to use the SAC model $use\_e$, learning rate $\eta$, task-shared parameters $\theta$, task-specific parameters $\left\lbrace \psi_i \right\rbrace^n_{i=1}$
		\Ensure updated task-shared parameters $\theta'$, updated task-specific parameters $\left\lbrace\psi'_i\right\rbrace^n_{i=1}$
		
		\If {$ce=2$ \textbf{and} $cb=bn$}
			\State Assign $\bm{L_{base}}$ to the average losses across every batch in the current epoch of all tasks;
		\EndIf
		
		\If{$ce > 2$}
            \State Add $(s_{t-1}, a_{t-1}, r_{t-1}, s_{t})$ into $buffer$;
		\EndIf
		
		\If{$ce \geq update\_e$}
			\State Train the SAC model $sac.\mathrm{train}(buffer)$;
		\EndIf
		
		\If{$ce < use\_e$}
			\State $\bm{\lambda_{v2}} = n \times \mathrm{softmax}(\mathrm{random\_normal}(n))$;
		\Else
			\State $\bm{\lambda_{v2}} = sac.\mathrm{select\_action}(\bm{L})$;
		\EndIf
        
        \State $\theta' = \theta - \eta \nabla_\theta \sum^n_{i=1}\lambda_{v2, i} \log(L_i)$;
		\For{i = 1 \textbf{to} n}
		\State $\psi_i' = \psi_i - \eta \nabla_{\psi_i} \lambda_{v2, i} \log(L_i)$;
		\EndFor
	\end{algorithmic}
\end{algorithm}

Unlike deploying DRL to play games \citep{mnih2013playing} which requires millions of episodes, only one complete training is allowed to obtain the optimal model in MTL. Therefore, in this work, many details are redesigned to accommodate the differences between these two different domains.
First, we consider the fundamental elements of DRL including environment, state, action, and reward:
\begin{itemize}
    \item \textbf{Environment:} We regard the entire MTL model and the training data for MTL as the environment.
    \item \textbf{State:} The state is required to properly describe the current environment and not be too high-dimensional for learning efficiency. Therefore we choose the losses of the current batch as the state $s_t$, which is determined by both the current parameter situation of the MTL model and the input training data of the current batch. In this way, the past training process of the MTL model can be fully represented by the current state, which can be formulated as a standard Markov decision process.
    \item \textbf{Action:} We regard the task weights of the current batch as the action $a_t$, and limit it to positive numbers that sum to $n$ for a fair comparison with other algorithms. 
    \item \textbf{Reward:} Inspired by maximizing the minimum improvement of all tasks in \citet{sener2018multi} and \citet{liu2021conflict}, we regard the minimum loss decline of all tasks in the current batch as the reward, where the  losses are also normalized by the average losses of the second epoch $\bm{L_{base}}$.
    Additionally, since reducing the losses of the MTL model is more challenging when the learning rate is lower than it was at the start of training, we add a multiplier factor $\alpha$ to the reward, which is calculated as the ratio of the initial learning rate to the current learning rate.
    Finally, the reward is defined as:
	\begin{equation}
        r_t = \alpha \times \min \left( \frac{\bm{L_t} - \bm{L_{t+1}}}{\bm{L_{base}}}\right),
	\end{equation}
	where $\bm{L_t}$ is the vector of the current batch losses and $\bm{L_{t+1}}$ is that of the next batch losses.
\end{itemize}

The SAC model adaptively assigns task weights to maximize the cumulative reward, so that the cumulative loss decline of each task can be maximized to the improvable gap. In other words, IGBv2 jointly minimizes the improvable gaps of all tasks by gradually minimizing the training losses to the desired final training losses. %through sequential decisions on task weights.

Furthermore, we present the training process of IGBv2 in Algorithm~\ref{alg:IGBv2}, where we initially set both the current epoch $cp$ and the current batch $cb$ to $1$.
At the beginning of training, since the SAC model is not sufficiently trained to be used, RLW \citep{lin2022reasonable} is deployed to randomly assign task weights for MTL, which is also beneficial for the training of the SAC model through random exploration.
After obtaining $\bm{L_{base}}$, we can get the state $s_t$, the action $a_t$, and the reward $r_{t-1}$ in each training batch of the MTL model. %After obtaining $\bm{L_{base}}$, we can get the state and action in the current batch, and get the reward in the next batch. 
These data are continually added into the replay buffer for training the SAC model.
Once the SAC model is trained well enough, it is deployed to assign task weights instead of RLW.

Additionally, we carefully redesign the replay buffer size, which is significantly smaller than typical SAC applications. 
When the buffer size is too large, since the performance of the MTL model is gradually improved and the training losses are reduced gradually, training the SAC model with too earlier historical data is not beneficial to the current training of the MTL model. Conversely, when the buffer size is too small, the training instability caused by time series data and low sample efficiency may make the training of the SAC model unsatisfactory, thereby also leading to poor performance of the MTL model.

\subsection{Combination of Loss Balancing and Gradient Balancing}
\label{sec:combination}

\begin{algorithm}[t!]
	\renewcommand{\algorithmicrequire}{\textbf{Input:}}
	\renewcommand{\algorithmicensure}{\textbf{Output:}}
	\caption{Training iteration of combining loss balancing and gradient balancing}  
	\label{alg:combination}
	\begin{algorithmic}[1]
		\Require loss balancing algorithm $LB$, gradient balancing algorithm $GB$, task number $n$, current batch losses $\bm{L}$, learning rate $\eta$, task-shared parameters $\theta$, task-specific parameters $\left\lbrace \psi_i \right\rbrace^n_{i=1}$
		\Ensure updated task-shared parameters $\theta'$, updated task-specific parameters $\left\lbrace\psi'_i\right\rbrace^n_{i=1}$
		
		\State Compute the task weights assigned by loss balancing $\bm{\lambda_l} = LB(\bm{L})$;
        \For{i = 1 \textbf{to} n}
            \State $g_i = \nabla_\theta \lambda_{l,i} \log(L_i)$;
        \EndFor
        \State $\theta' = \theta - \eta GB(g_1, \cdots, g_n)$;
		\For{i = 1 \textbf{to} n}
		  \State $\psi_i' = \psi_i - \eta \nabla_{\psi_i} \lambda_{l, i} \log(L_i)$;
		\EndFor
		
	\end{algorithmic}
\end{algorithm}

Our IGB algorithms dynamically provide varying importance for each task to balance the improvable gaps, while gradient balancing algorithms deal more directly with gradient conflicts at the gradient level. These two types of algorithms are complementary and can be combined together for further improvements. 

We first assign task weights with the loss balancing algorithm and then input the weighted losses to the gradient balancing algorithm to obtain the final update gradient. In this way, the performance can be further improved while keeping the training time almost the same as that of gradient balancing alone, since most loss balancing algorithms hardly add extra training time.

Typically, updating the model with existing gradient balancing algorithms can be divided into two ways: one assigns weights to gradients of both task-shared and task-specific parameters, while the other only aggregates gradients of task-shared parameters. As illustrated in Algorithm~\ref{alg:combination}, when combining loss balancing and gradient balancing in the latter case, which is more common, task-shared parameters are updated by both loss balancing and gradient balancing, while task-specific parameters are updated independently by loss balancing.

\begin{table*}[t]
    \centering	
    \caption{Comparative results on the NYUv2 dataset for multi-task scene understanding. $\uparrow (\downarrow)$ indicates that the higher (lower) the result, the better the performance. $+$ represents the combination of loss balancing and gradient balancing. In each group, the best results are \textbf{bolded} and the second-best results are \underline{underlined}.}
    \label{tab:nyu}
    \scalebox{0.90}{
    \tabcolsep4pt
	\begin{tabular}{lccccccccccc}
		\toprule[1.5pt] % from booktabs package
		\multirow{3}{*}{Methods} & \multicolumn{2}{c}{Segmentation} & \multicolumn{2}{c}{Depth} & \multicolumn{5}{c}{Surface Normal} & \multirow{3}{*}{$\bm{\Delta m\downarrow}$} & \multirow{3}{*}{$\bm{T\downarrow}$} \\
		%\cmidrule{2-3} \cmidrule{4-5} \cmidrule{6-10}
		\multirow{3}{*}{ } &  \multirow{2}{*}{mIoU$\uparrow$} & \multirow{2}{*}{Pix Acc$\uparrow$} & \multirow{2}{*}{Abs Err$\downarrow$} & \multirow{2}{*}{Rel Err$\downarrow$} & \multicolumn{2}{c}{Angle Distance$\downarrow$}& \multicolumn{3}{c}{Within $t^\circ$$\uparrow$} & \multirow{3}{*}{ } & \multirow{3}{*}{ } \\
		%\cmidrule{6-7} \cmidrule{8-10}
		\multirow{3}{*}{ } &  \multirow{2}{*}{ } & \multirow{2}{*}{ } & \multirow{2}{*}{ } & \multirow{2}{*}{ } & Mean & Median & 11.25 & 22.5 & 30 & \multirow{3}{*}{ } & \multirow{3}{*}{ } \\
		\midrule % from booktabs package
		STL & 41.16 & 65.70 & 0.6074 & 0.2400 & 24.49 & 18.24 & 31.92 & 59.16 & 70.56 & & \\
		\midrule
		EW & 40.12	& 66.16	& 0.5189 & 0.2039 & 28.30 &	23.58 &	23.07 &	48.35 &	61.39 &	8.45 & 1.00\\
		RLW	\citep{lin2022reasonable}& 39.72 & 65.40 & 0.5252 & 0.2092 & 28.85 & 24.38 & 21.59 & 46.74 & 59.98 & 10.82 &  1.00\\
		DWA	\citep{liu2019end} & \underline{41.60} & \underline{66.52} &	\underline{0.5041} & \textbf{0.2000} & 28.11	& 23.32	& 23.32	& 48.71 & 61.83 & 7.08 & 1.00\\
		UW \citep{kendall2018multi} & 40.29	& 64.60 & 0.5081 & 0.2058 & 26.69 & 21.47 & 26.18 & 52.43 & 65.21 & 4.09 &  1.00\\
		IGBv1~(ours) & 39.91 & 66.03 & \textbf{0.4961} & \underline{0.2009} & \underline{26.08} & \underline{20.69} & \underline{27.52} & \underline{54.11} & \underline{66.57} & \underline{1.76} &  1.00\\
		IGBv2~(ours) & \textbf{41.66} & \textbf{66.98} & 0.5392 & 0.2090 & \textbf{25.47} & \textbf{19.88} & \textbf{28.79} & \textbf{55.86} & \textbf{68.14} & \textbf{0.50} & 1.09\\
		\midrule
		MGDA \citep{sener2018multi} & 30.55 & 60.20 & 0.5728 & 0.2179 & \textbf{24.16} & \textbf{18.12} & \textbf{32.34} & \textbf{59.51} &	\textbf{71.04} & 1.63 & 2.88\\
		PCGrad \citep{yu2020gradient} & 42.08 & 67.18	& 0.5098 & 0.1994 & 27.48 & 22.46 & 24.79 & 50.46 &	63.42 & 5.01 & 2.52\\
		CAGrad \citep{liu2021conflict} & 41.43 & 67.35 & 0.5042 & 0.1976 & 25.25 & 19.58 & 29.31 & 56.43 & 68.58 & -1.31 & 2.85\\
		IMTL-G \citep{liu2021towards} & 41.67	& 67.18	& 0.5019 & 0.1977 &	25.03 & 19.47 & 29.61 & 56.70 & 68.95 & -1.76 & 2.85\\
		Nash \citep{navon2022multi}& \underline{42.96} & \underline{68.30} &	0.4966 & 0.1986 & 24.79 & 18.97 & 30.50 & 57.74 & 69.67 & -3.39 & 3.16\\
		IGBv1 + MGDA & 40.93 & 66.66 & 0.5238 & 0.2049 & \underline{24.38} & \underline{18.44} & \underline{31.52} & \underline{58.97} & \underline{70.66} & -3.03 & 2.90\\
		IGBv2 + MGDA & 40.91 & 67.01 & 0.5359 & 0.2060 & 25.11 & 19.41 & 29.65 & 56.77 & 68.89 & -0.54 & 2.96\\
		IGBv1 + PCGrad & 40.57 & 66.98 & 0.4989 & 0.1990 & 25.76 & 20.33 & 28.12& 54.80 & 67.30 & 0.56 & 2.52\\
		IGBv2 + PCGrad & 36.00 & 63.04 & 0.5286 & 0.2079 & 26.00 & 20.41 & 28.11 & 54.65 & 66.96 & 3.66 & 2.60\\
		IGBv1 + CAGrad & 40.98 & 66.84 & 0.5050 & 0.1995 & 25.15 & 19.49	& 29.54 & 56.57 & 68.74 & -1.23 & 2.86\\
		IGBv2 + CAGrad & 40.12 & 66.23 & 0.5190 & 0.2004 & 24.69 & 18.77 & 30.72 & 58.19 & 70.00 & -2.16 & 2.93\\
		
		IGBv1 + IMTL-G & 42.01 & 67.69 & 0.5009 & 0.1995 & 24.97 & 19.21 & 30.01 & 57.20	& 69.28 & -2.35 & 2.85\\
		IGBv2 + IMTL-G & 41.25 & 66.43 & 0.5134 & 0.2011 & 25.18	& 19.58	& 29.17 & 56.53 & 68.77	& -0.81 & 2.94\\
		IGBv1 + Nash & 42.83 & 67.99 & \textbf{0.4892} & \textbf{0.1958} & 24.70 &	18.92 & 30.64 & 57.79 & 69.78 & \underline{-3.71} & 3.17\\
		IGBv2 + Nash & \textbf{43.97} & \textbf{68.53} & \underline{0.4928} & \underline{0.1967} & 24.71 & 18.83 & 30.83 & 57.90 &	69.86 & \textbf{-4.15} & 3.23\\
		\bottomrule[1.5pt] % from booktabs package
	\end{tabular}}
\end{table*}

\section{Experiments}

\subsection{Experimental Setup}

\noindent\textbf{Datasets.}~~
We train and evaluate our model on the NYUv2 dataset \citep{silberman2012indoor} for multi-task scene understanding and on the QM9 dataset \citep{ramakrishnan2014quantum} for multi-task regression prediction. 

\textbf{NYUv2} is an indoor scene image dataset with dense pixel-level 13-class labeling. It comprises 795 training samples and 654 test samples. Prior studies \citep{liu2021conflict, navon2022multi} typically regard the average performance of the last 10 epochs on the test set as the final result, which is unreasonable in machine learning research. Therefore, we randomly divide the 654 original test samples into 197 validation samples (the validation set) and 457 test samples (the test set).

\textbf{QM9} is a chemical molecule dataset widely used for graph neural networks (GNNs) \citep{wu2020comprehensive}, with approximately 130K molecular samples (which are represented as graphs annotated with node and edge features). Following \citet{navon2022multi}, we use this dataset for regression prediction on 11 properties of chemical molecules, and we use 110K samples for training, 10K samples for validation, and 10K samples for testing.

\noindent\textbf{Compared Methods.}~~
We compare our methods with the following classic algorithms that have been described in Section~\ref{sec:related}: 
(1) Equal Weighting (EW) which minimizes $\sum_{i=1}^{n} L_i$;
(2) Random Loss Weighting (RLW) \citep{lin2022reasonable};
(3) Dynamic Weight Average (DWA) \citep{liu2019end};
(4) Uncertainty Weighting (UW) \citep{kendall2018multi};
(5) Multiple Gradient Descent Algorithm (MGDA) \citep{sener2018multi};
(6) Projecting Conflicting Gradient (PCGrad) \citep{yu2020gradient};
(7) Conflict-Averse Gradient (CAGrad) \citep{liu2021conflict};
(8) Impartial Multi-Task Learning (IMTL-G) \citep{liu2021towards};
(9) Nash-MTL (Nash) \citep{navon2022multi}.

\noindent\textbf{Evaluation Metrics.}~~
In the experiments, we first report the common evaluation metrics for each task over each dataset. Moreover, since the significance of MTL optimization is to improve both model performance and learning efficiency, we report two overall metrics to comprehensively evaluate MTL optimization methods:

(1) $\Delta m$: the average per-task performance drop compared with STL, which is defined as:
\begin{equation}
	\Delta m = \frac{1}{K} \sum_{k=1}^{K}(-1)^{\delta_k}\frac{M_{m,k} - M_{b,k}}{M_{b,k}},
\end{equation}
where $K$ is the total number of common evaluation metrics over all tasks, $M_{m,k}$ is the value on the $k$-th metric of the evaluated method and $M_{b,k}$ is that of the STL baseline, $\delta_k = 1$ if a higher value is better for the $k$-th metric and $0$ otherwise \citep{liu2021conflict, navon2022multi}.

(2) $T$: the relative training time compared with EW, which is calculated as the ratio of the training time of the evaluated method to that of the EW baseline.

\begin{figure*}
    \centering
    \includegraphics[width=\linewidth]{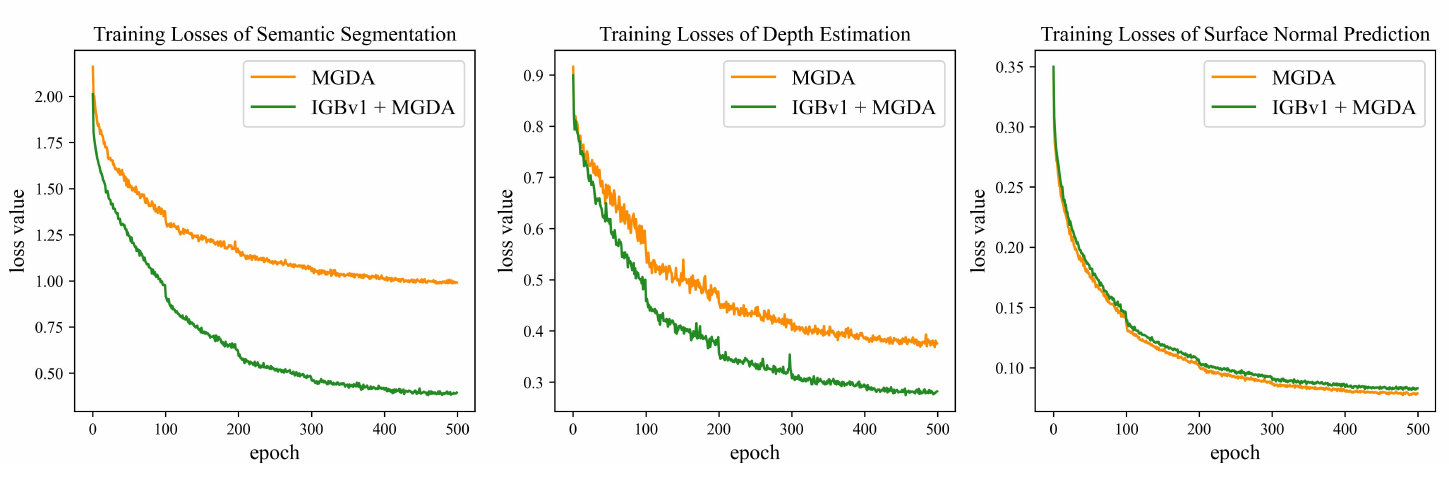}
    \caption{Comparison of loss decline curves for individual tasks of MGDA and IGBv1~+~MGDA.}
    \label{fig:mgda}
\end{figure*}

\noindent\textbf{Implementation Details.}~~
In the experiments on the NYUv2 dataset, we train a SegNet model from \citet{badrinarayanan2017segnet} for 500 epochs with the Adam optimizer \citep{kingma2014adam}, while the learning rate is initially set to $1e\text{-}4$ and decays by half for every 100 epochs. Then we test the model which has the best \emph{validation} metric $\Delta m$ to obtain the final algorithm performance.

In the experiments on the QM9 dataset, following \citet{navon2022multi}, we normalize each task target to have zero mean and unit standard deviation, and use the code implemented by \citet{fey2019fast}, which contains a popular GNN from \citet{gilmer2017neural} and a pooling operator from \citet{vinyals2015order}. 
We train the model for 300 epochs with the Adam optimizer \citep{kingma2014adam}, while the learning rate is searched in $\left\lbrace 1e\text{-}3, 5e\text{-}4, 1e\text{-}4 \right\rbrace$ and decreased by the ReduceOnPlateau scheduler according to the \emph{validation} metric $\Delta m$.

For the IGBv2 algorithm, the SAC model is updated once every 50 batches during the MTL model training, where the learning rate of the SAC model is adjusted according to the situation: $1e\text{-}4$ when combining IGBv2 with MGDA \citep{sener2018multi}, PCGrad \citep{yu2020gradient} or Nash \citep{navon2022multi}, and $3e\text{-}4$ when combining IGBv2 with other gradient balancing algorithms or using IGBv2 alone.
The replay buffer size is set to $1e4$. The discount factor $\gamma$ is set to $0.99$. The start epoch to update the SAC model $update\_e$ is set to $4$, and the start epoch to use the SAC model $use\_e$ is set to $6$.

\subsection{Results of Multi-Task Scene Understanding on NYUv2}

Scene understanding on the NYUv2 dataset is the most common evaluation scenario in the MTL research, which contains three tasks: semantic segmentation, depth estimation, and surface normal prediction.

The comparative results are shown in Table~\ref{tab:nyu}. To verify the complementarity between IGB and gradient balancing, we combine our IGB algorithms with all compared gradient balancing algorithms.
We categorize all methods into two groups according to efficiency and compare the performance in each group: one is the loss balancing algorithms; the other is the gradient balancing algorithms and also the combined algorithms by IGB and gradient balancing.

First, we focus on the overall performance and learning efficiency of MTL.
\textbf{In terms of overall performance ($\bm{\Delta m}$):} 
Our IGB algorithms are the best in loss balancing and also competitive with some gradient balancing algorithms. 
Combining IGB and gradient balancing is better than using gradient balancing alone. 
In particular, IGBv1~+~Nash and IGBv2~+~Nash achieve better performance than Nash \citep{navon2022multi}, yielding the SOTA performance of all algorithms.
\textbf{In terms of learning efficiency ($\bm{T}$):}
The training time of gradient balancing is about three times that of loss balancing.
Either used alone or combined with gradient balancing, IGBv1 hardly leads to extra training time, and IGBv2 slightly leads to extra training time but can achieve better performance than IGBv1 in most situations.

Notably, with our IGB algorithms, users now have more freedom to take the trade-off between efficiency and performance in real-world MTL applications: our IGB algorithms could be chosen if efficiency is more important, and IGB~+~gradient balancing algorithms could be chosen if performance is more important. Specifically, IGBv1 is more efficient and simpler to implement, while IGBv2 achieves better performance in most situations.

Next, we analyze the superiority of our IGB algorithms through the specific task performance. Compared with STL, the performance of most existing algorithms is close to or even better on semantic segmentation and depth estimation, but significantly worse on surface normal prediction. To some extent, it means that in the NYUv2 MTL scenario, semantic segmentation and depth estimation are simpler, while surface normal prediction is more difficult. Such performance imbalance is undesirable in most MTL scenarios. As expected, our IGB algorithms can effectively alleviate this imbalance problem. That is, compared with other loss balancing algorithms, the performance of our IGBv1 and IGBv2 is competitive on semantic segmentation and depth estimation, and significantly better on surface normal prediction, which is close to that of STL.  

Moreover, the performance of MGDA \citep{sener2018multi} is the best on surface normal prediction among all algorithms, but significantly worse than other algorithms on semantic segmentation and depth estimation, leading to its poor overall performance. By combining MGDA and our IGB algorithms, we significantly alleviate the performance imbalance problem and greatly improve its overall performance. 
As shown in Figure~\ref{fig:mgda}, when using IGBv1, the training losses of semantic segmentation and depth estimation decrease faster and are significantly lower. Meanwhile, for surface normal prediction, the training loss of IGBv1~+~MGDA is comparable to that of MGDA during the training process.
The main reason is that our improvable gap balancing algorithms can automatically focus on tasks that have not been properly trained.

\subsection{Results of Multi-Task Regression Prediction on QM9}

Predicting 11 properties of chemical molecules on the QM9 dataset is a more challenging MTL problem because of the larger task number, lower task relevance, and more varied task difficulty. Prior studies have found that the performance of STL on 11 property prediction is significantly better than that of MTL \citep{maron2019provably, gasteiger2020directional}, but driven by the need for higher learning efficiency and smaller model size, MTL methods are still worth studying.

The comparative results on the QM9 dataset are shown in Table~\ref{tab:qm}. We can see that our IGB algorithms beat all compared algorithms except Nash \citep{navon2022multi} in performance, and perform much better than all gradient balancing algorithms in learning efficiency. For example, the training time of Nash is about seven times that of our IGB algorithms with close performance, and the training time of MGDA \citep{sener2018multi} is twenty-two times that of our IGB algorithms due to aggregating the high-dimensional gradients by multi-objective optimization.
As the number of tasks increases, the learning efficiency gap between loss balancing and gradient balancing becomes larger and larger. In a scenario like QM9, loss balancing (especially IGB) is thus more valuable.
Unfortunately, the weighting or aggregating decisions made by existing loss balancing and gradient balancing algorithms are both unsatisfactory on this problem (much worse than STL). Combining loss balancing and gradient balancing is equivalent to combining the decisions made by the two types of algorithms, which has not yielded better performance.

\subsection{Ablation Study}

As shown in Table~\ref{tab:ablation}, we analyze the contributions of improvable gap balancing in our algorithms and the effects of different settings in the IGBv2 algorithm on the NYUv2 dataset. 
The compared methods include: 
(1) EW which minimizes $\sum^n_{i=1}L_i$; 
(2) SI which minimizes $\sum^n_{i=1}\log (L_i)$ \citep{navon2022multi};
(3) IGBv1 (full); 
(4) IGBv2 (reward / min, buffer $1e4$) where the reward is calculated by the average loss decline instead of the minimum; 
(5) IGBv2 (reward / $\alpha$, buffer~$1e4$) where the reward is calculated without the multiplier factor $\alpha$; 
(6) IGBv2 (full reward, buffer $5e3 / 1e4 / 5e4 / 1e6$) where the replay buffer size is set to $5e3$, $1e4$, $5e4$, or $1e6$, and IGBv2 (full reward, buffer $1e4$) is our full IGBv2.

\begin{table*}[t]
    \centering
    \caption{Comparative results on the QM9 dataset for multi-task regression prediction. $\uparrow (\downarrow)$ indicates that the higher (lower) the result, the better the performance. In each group, the best results are \textbf{bolded} and the second-best results are \underline{underlined}.}
    \label{tab:qm}
    \scalebox{0.90}{
    \tabcolsep3pt
    \begin{tabular}{lccccccccccccc}
        \toprule[1.5pt] % from booktabs package
        \multirow{2}{*}{Methods}  & $\mu$ & $\alpha$ & $\epsilon_\mathrm{HOMO}$ & $\epsilon_\mathrm{LUMO}$ & $\left\langle R^2 \right\rangle$ & ZPVE & $U_0$ & $U$ & $H$ & $G$ & $c_v$ & \multirow{2}{*}{$\bm{\Delta m\downarrow}$} & \multirow{2}{*}{$\bm{T\downarrow}$} \\
        \multirow{2}{*}{ } & \multicolumn{11}{c}{MAE$\downarrow$} & \multirow{2}{*}{ } & \multirow{2}{*}{ }\\
        \midrule % from booktabs package
        STL & 0.067 & 0.181 & 60.57 & 53.91 & 0.502 & 4.53 & 58.8 & 64.2 & 63.8 & 66.2 & 0.072\\
        \midrule % from booktabs package
        EW	& \textbf{0.106} & \textbf{0.325} & \textbf{73.57} & \textbf{89.67} & 5.19 & 14.06 & 143.4 & 144.2 & 144.6 & 140.3 & 0.128 & 177.6 & 1.00\\
        RLW	\citep{lin2022reasonable}& 0.113 & 0.340 & 76.95 & 92.76 & 5.86 & 15.46 & 156.3 & 157.1 & 157.6 & 153.0 & 0.137 & 203.8 & 1.00\\
        DWA	\citep{liu2019end} & \underline{0.107} & \textbf{0.325} & \underline{74.06} & \underline{90.61} & 5.09 & 13.99 & 142.3 & 143.0 & 143.4 & 139.3 & 0.125 & 175.3 & 1.00\\
        UW \citep{kendall2018multi}	& 0.386 & 0.425 & 166.2 & 155.8 & \underline{1.06} & 4.99 & 66.40 & 66.78 & 66.80 & 66.24 & 0.122 & 108.0 & 1.02\\
        IGBv1~(ours)	& 0.271	& 0.351 & 143.1 & 132.4 & 1.07 & \underline{4.53} & \textbf{56.74} & \textbf{57.13} & \textbf{57.16} & \textbf{56.73} & \underline{0.115} & \underline{73.9} & 1.01\\
        IGBv2~(ours)	& 0.251	& 0.333 & 149.1 & 130.2	& \textbf{0.956} & \textbf{4.39} & \underline{56.75} & \underline{57.19} & \underline{57.25} & \textbf{56.73} & \textbf{0.110} & \textbf{67.7} & 1.16\\ \hline
        MGDA \citep{sener2018multi} & 0.217 & 0.368 & 126.8 & 104.6 & 3.22 & \underline{5.69} & 88.37 & \underline{89.40} & \underline{89.32} & \underline{88.01} & 0.120 & 120.5 & 22.90\\
        PCGrad \citep{yu2020gradient} & \underline{0.106} & 0.293 & \textbf{75.85} & \underline{88.33} & 3.94 & 9.15 & 116.4 & 116.8 & 117.2 & 114.5 & 0.110 & 125.7 & 5.85\\
        CAGrad \citep{liu2021conflict} & 0.118 & 0.321 & 83.51 & 94.81 & 3.21 & 6.93 & 114.0 & 114.3 & 114.5 & 112.3 & 0.116 & 112.8 & 4.97\\
        IMTL-G \citep{liu2021towards} & 0.136 & \underline{0.287} & 98.31 & 93.96 & \textbf{1.75} & \underline{5.69} & \underline{101.4} & 102.4 & 102.0 & 100.1 & \underline{0.096} & \underline{77.2} & 4.96\\
        Nash \citep{navon2022multi} & \textbf{0.102} & \textbf{0.248} & \underline{82.95} & \textbf{81.89} & \underline{2.42} & \textbf{5.38} & \textbf{74.5} & \textbf{75.02} & \textbf{75.10} & \textbf{74.16} & \textbf{0.093} & \textbf{62.0} & 7.20\\
        \bottomrule[1.5pt] % from booktabs package
    \end{tabular}}
\end{table*}

\begin{table}[t!]
    \centering
    \caption{Results of ablation study on the NYUv2 dataset. `reward~/~min' (or  `reward~/~$\alpha$') denotes that min (or $\alpha$) is removed from the definition of the full reward. }
    \label{tab:ablation}
    \scalebox{0.90}{
    \tabcolsep11pt
	\begin{tabular}{lcc}
		\toprule[1.5pt] % from booktabs package
		Methods  & $\bm{\Delta m\downarrow}$ & $\bm{T\downarrow}$ \\
		\midrule % from booktabs package
        EW & 8.45 & 1.00\\
        SI \citep{navon2022multi} & 2.36 & 1.00\\
		IGBv1~(full) & 1.76 & 1.00\\
        IGBv2~(reward~/~min, buffer~$1e4$) & 5.97 & 1.09\\
        IGBv2~(reward~/~$\alpha$, buffer~$1e4$) & 0.91 & 1.09\\
		IGBv2~(full reward, buffer~$5e3$) & 4.78 & 1.09\\
        IGBv2~(full reward, buffer~$1e4$) & \textbf{0.50} & 1.09\\
        IGBv2~(full reward, buffer~$5e4$) & 2.72 & 1.09\\
        IGBv2~(full reward, buffer~$1e6$) & 3.48 & 1.06\\
		\bottomrule[1.5pt] 
	\end{tabular}}
\end{table}

We can observe from Table~\ref{tab:ablation} that: 
(1) SI brings significant performance improvement compared with EW, indicating that the scale-invariant loss balancing paradigm is superior to the traditional loss balancing paradigm.
(2) Our full IGBv1 and IGBv2 further improve the performance compared with SI, which clearly validates the effectiveness of improvable gap balancing.
(3) IGBv2 (reward / min, buffer $1e4$), IGBv2 (reward / $\alpha$, buffer~$1e4$), and IGBv2 (full reward, buffer $5e3 / 5e4 / 1e6$) all produce performance degradation compared with the full IGBv2, indicating that our redesign of the reward and the replay buffer size is necessary when deploying DRL for MTL optimization.

Moreover, as shown in Table~\ref{tab:ablation2}, we combine existing loss balancing methods with the SI paradigm (UW~+~SI method does not work since UW may not be compatible with SI), and their performance on NYUv2 is much lower than that of our IGB methods. It provides further evidence of the effectiveness of our IGB methods, i.e., the task weighting of IGB and the SI paradigm are complementary (combining them leads to better performance).

\begin{table}[t!]
    \centering
    \caption{Results of loss balancing using the SI paradigm on the NYUv2 dataset. $+$ represents the combination of a loss balancing method and the SI paradigm.}
    \label{tab:ablation2}
    \scalebox{0.90}{
    \tabcolsep17pt
	\begin{tabular}{lcc}
		\toprule[1.5pt] % from booktabs package
		Methods  & $\bm{\Delta m\downarrow}$ & $\bm{T\downarrow}$ \\
		\midrule % from booktabs package
        SI \citep{navon2022multi} & 2.36 & 1.00\\
        RLW \citep{lin2022reasonable} & 10.82 & 1.00\\
        RLW + SI & 6.24 & 1.00\\
        DWA \citep{liu2019end} & 7.08 & 1.00\\
        DWA + SI & 3.25 & 1.00\\
        UW \citep{kendall2018multi} & 4.09 & 1.00\\
        UW + SI & - & -\\
        IGBv1 & 1.76 & 1.00\\
        IGBv2 & \textbf{0.50} & 1.09\\
		\bottomrule[1.5pt] 
	\end{tabular}}
\end{table}

\section{Conclusion}

In this paper, we propose two novel loss balancing algorithms, IGBv1 and IGBv2. 
We dynamically assign task weights to balance the improvable gaps: IGBv1 takes a simple heuristic, and IGBv2 (for the first time) applies DRL to MTL. 
We analyze the complementarity between IGB and gradient balancing. Extensive experiments show that our IGB algorithms outperform all existing loss balancing algorithms and bring further performance improvements when combined with gradient balancing. 
More importantly, we reemphasize the significance of loss balancing in MTL in terms of learning efficiency and give users more freedom to take the trade-off between efficiency and performance in real-world applications.

\begin{acknowledgements} 
This work was supported in part by National Natural Science Foundation of China (61976220), and the Fundamental Research Funds for the Central Universities and the Research Funds of Renmin University of China (23XNH026). Zhiwu~Lu is the corresponding author.
\end{acknowledgements}

% References
\bibliography{dai_520}
\end{document}